\documentclass[journal]{IEEEtran}
\usepackage{graphicx}
\usepackage{cite}
\usepackage{picinpar}
\usepackage{amsmath}
\usepackage{stfloats}
\usepackage{url}
\usepackage{flushend}
\usepackage[latin1]{inputenc}
\usepackage{colortbl}
\usepackage{soul}
\usepackage{multirow}
\usepackage{pifont}
\usepackage{color}
\usepackage{alltt}
\usepackage[hidelinks]{hyperref}
\usepackage{enumerate}
\usepackage{siunitx}
\usepackage{breakurl}
\usepackage{epstopdf}
\usepackage{pbox}
\usepackage{algorithm}
\usepackage{algpseudocode}
\usepackage{subfigure}

\newcommand{\circR}{\operatornamewithlimits{circ}}

\begin{document}
\title{A Survey of Model Compression and Acceleration for Deep Neural Networks}

\author{Yu~Cheng, Duo~Wang, 
        Pan~Zhou,~\IEEEmembership{Member,~IEEE,}
        and~Tao~Zhang,~\IEEEmembership{Senior Member,~IEEE}

\thanks{Yu Cheng is a Senior Researcher at Microsoft, One Microsoft Way, Redmond, WA 98052, USA.}
\thanks{Duo Wang and Tao Zhang are with the Department of Automation, Tsinghua University, Beijing 100084, China.}
\thanks{Pan Zhou is with the School of Electronic Information and Communications, Huazhong University of Science and Technology, Wuhan 430074, China.}}

\markboth{IEEE SIGNAL PROCESSING MAGAZINE, SPECIAL ISSUE ON DEEP LEARNING FOR IMAGE UNDERSTANDING}%
{Shell \MakeLowercase{\textit{et al.}}: Bare Demo of IEEEtran.cls for IEEE Journals}

\maketitle
	
\begin{abstract}
Deep neural networks (DNNs) have recently achieved great success in many visual recognition tasks. However, existing deep neural network models are computationally expensive and memory intensive, hindering their deployment in devices with low memory resources or in applications with strict latency requirements. Therefore, a natural thought is to perform model compression and acceleration in deep networks without significantly decreasing the model performance. During the past five years, tremendous progress has been made in this area. In this paper, we review the recent techniques for compacting and accelerating DNN models. In general, these techniques are divided into four categories: parameter pruning and quantization, low-rank factorization, transferred/compact convolutional filters, and knowledge distillation. Methods of parameter pruning and quantization are described first, after that the other techniques are introduced. For each category, we also provide insightful analysis about the performance, related applications, advantages and drawbacks. Then we go through some very recent successful methods, for example, dynamic capacity networks and stochastic depths networks. After that, we survey the evaluation matrices, the main datasets used for evaluating the model performance and recent benchmark efforts. Finally, we conclude this paper, discuss remaining the challenges and possible directions for future work. 
\end{abstract}


\IEEEpeerreviewmaketitle

\section{Introduction}
In recent years, deep neural networks have recently received lots of attention, been applied to different applications and achieved dramatic accuracy improvements in many tasks. These works rely on deep networks with millions or even billions of parameters, and the availability of GPUs with very high computation capability plays a key role in their success. For example, the work by Krizhevsky \emph{et al.} \cite{krizhevsky2012} achieved breakthrough results in the 2012 ImageNet Challenge using a network containing 60 million parameters with five convolutional layers and three fully-connected layers. Usually, it takes two to three days to train the whole model on ImagetNet dataset with a NVIDIA K40 machine. Another example is the top face verification results on the Labeled Faces in the Wild (LFW) dataset were obtained with networks containing hundreds of millions of parameters, using a mix of convolutional, locally-connected, and fully-connected layers \cite{yaniv2014,DBLP:journals/corr/LuKZCJF16}. It is also very time-consuming to train such a model to get reasonable performance. In architectures that rely only on fully-connected layers, the number of parameters can grow to billions \cite{dean2012}.

As larger neural networks with more layers and nodes are considered, reducing their storage and computational cost becomes critical, especially for some real-time applications such as online learning and incremental learning. In addition, recent years witnessed significant progress in virtual reality, augmented reality, and smart wearable devices, creating unprecedented opportunities for researchers to tackle fundamental challenges in deploying deep learning systems to portable devices with limited resources (e.g. memory, CPU, energy, bandwidth). Efficient deep learning methods can have significant impacts on distributed systems, embedded devices, and FPGAs for Artificial Intelligence. For example, the ResNet-50 \cite{DBLP:journals/corr/HeZRS15} with 50 convolutional layers needs over 95MB memory for storage and over 3.8 billion floating number multiplications when processing an image. After discarding some redundant weights, the network still works as usual but saves more than 75\% of parameters and 50\% computational time. For devices like cell phones and FPGAs with only several megabyte resources, how to compact the models used on them is also important.

Achieving these goals calls for joint solutions from many disciplines, including but not limited to machine learning, optimization, computer architecture, signal processing, and hardware design. In this paper, we review recent works on compressing and accelerating deep neural networks, which attracts a lot of attention from the deep learning community and already achieved lots of progress in the past years. 

Based on their properties, we divide these approaches into four categories: parameter pruning and quantization, low-rank factorization, transferred/compact convolutional filters, and knowledge distillation. The parameter pruning and quantization based methods explore the redundancy in the model parameters and try to remove the redundant and uncritical ones. Low-rank factorization based techniques use matrix/tensor decomposition to estimate the informative parameters of the DNNs. The approaches based on transferred/compact convolutional filters design special structural convolutional filters to reduce the parameter space and save storage/computation. The knowledge distillation based methods learn a distilled model and train a more compact neural network to reproduce the output of a larger network. 

In Table \ref{tab:all:sum}, we briefly summarize these four types of approaches. The parameter pruning \& quantization, low-rank factorization and knowledge distillation approaches can be deployed in DNN models with fully connected layers and convolutional layers, achieving comparable performances. On the other hand, methods using transferred/compact filters are designed for convolutional layers only. Low-rank factorization and transferred/compact filters based approaches provide an end-to-end pipeline and can be easily implemented in CPU/GPU environment. Parameter pruning \& quantization use different strategies such as binary coding and sparse constraints to perform the task.

Regarding the training protocols, models based on parameter pruning/quantization and low-rank factorization can be extracted from pre-trained models or trained from scratch. While the transferred/compact filter and knowledge distillation models can only support training from scratch. Most of these methods are independently designed and complementary to each other. For example, transferred layers and parameter pruning \& quantization can be deployed together. Another example is that, model quantization \& binarization can be used together with low-rank approximations to achieve further compression/speedup. We will describe the details of their properties, and analysis of strengths and drawbacks in the following sections separately.

\begin{table*}
\caption{Summarization of different approaches for model compression and acceleration.}
\label{tab:all:sum}
\centering
\begin{tabular}{|c|c|c|c|}
\hline Category Name & Description & Applications & More details \\
\hline Parameter pruning and quantization  &  Reducing redundant parameters which & Convolutional layer and  & Robust to various settings, can achieve \\
& are not sensitive to the performance & fully connected layer & good performance, can support both train  \\
&  &  & from scratch and pre-trained model \\
\hline Low-rank factorization & Using matrix/tensor decomposition to  & Convolutional layer and & Standardized pipeline, easily to be \\
& estimate the informative parameters & fully connected layer & implemented, can support both train \\
&  &  & from scratch and pre-trained model \\
\hline Transferred/compact convolutional &  Designing special structural convolutional &  Convolutional layer & Algorithms are dependent on applications, \\
filters & filters to save parameters & only & usually achieve good performance, \\
&  &  & only support train from scratch \\
\hline Knowledge distillation & Training a compact neural network with & Convolutional layer and  & Model performances are sensitive  \\
& distilled knowledge of a large model & fully connected layer & to applications and network structure\\
&  &  & only support train from scratch \\
\hline
\end{tabular}
\end{table*}

\section{Parameter Pruning and Quantization}
Early works showed that network pruning and quantization are effective in reducing the network complexity and addressing the over-fitting problem \cite{DBLP:journals/corr/GongLYB14}. After found that pruning can bring regularization to neural networks and hence improve generalization, it has been widely studied to compress DNNs. These techniques can be further mapped into three sub-categories: quantization and binarization, network pruning, and structural matrix. 

\subsection{Quantization and Binarization}
Network quantization compresses the original network by reducing the number of bits required to represent each weight. Gong \textit{et al.} \cite{DBLP:journals/corr/GongLYB14} and Wu et al. \cite{wu2016quantized} applied $k$-means scalar quantization to the parameter values. Vanhoucke \textit{et al.} \cite{37631} showed that 8-bit quantization of the parameters can result in significant speed-up with minimal loss of accuracy. The work in \cite{Gupta:2015:DLL:3045118.3045303} used 16-bit fixed-point representation in stochastic rounding based CNN training, which significantly reduced memory usage and float point operations with little loss in classification accuracy. 

\begin{figure}
\centering
\includegraphics[width=9cm]{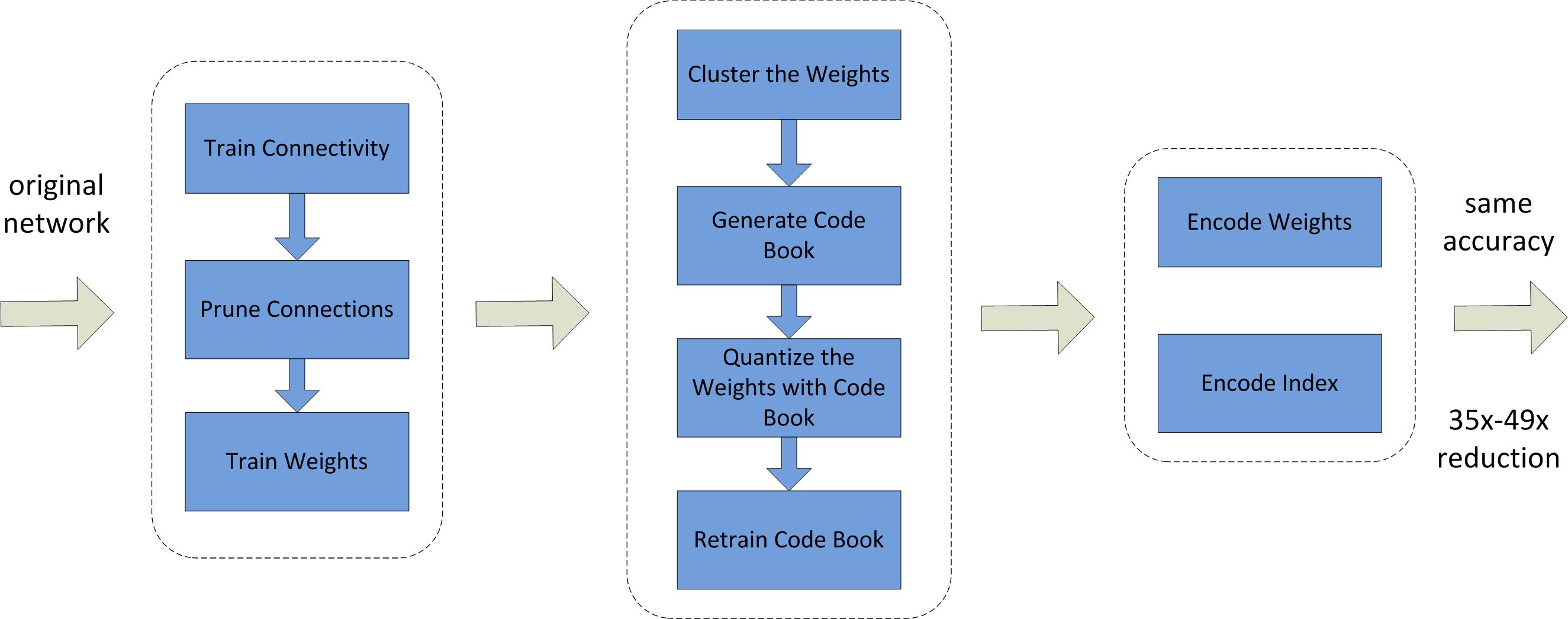}
\caption{The three-stage compression methods proposed in [10]: pruning, quantization and huffman encoding. The input is the original model and the output is the compressed model.}
\label{fig:quantization}
\end{figure}

The method proposed in \cite{han2015deep_compression} quantized the link weights using weight sharing and then applied Huffman coding to the quantized weights as well as the codebook to further reduce the rate. As shown in Figure \ref{fig:quantization}, it started by learning the connectivity via normal network training, followed by pruning the small-weight connections. Finally, the network was retrained to learn the final weights for the remaining sparse connections. This work achieved the state-of-art performance among all quantization based methods. In \cite{DBLP:journals/corr/ChoiEL16}, it was shown that Hessian weight could be used to measure the importance of network parameters, and proposed to minimize Hessian-weighted quantization errors in average to cluster parameters. Quantization is a very effective way for model compression and acceleration.

In the extreme case of the 1-bit representation of each weight, that is binary weight neural networks. The main idea is to directly learn binary weights or activation during the model training. There are several works that directly train CNNs with binary weights, for instance, BinaryConnect \cite{DBLP:conf/nips/CourbariauxBD15}, BinaryNet \cite{DBLP:journals/corr/CourbariauxB16} and XNOR \cite{rastegariECCV16}. A systematic study in \cite{DBLP:journals/corr/MerollaAAEM16} showed networks trained with back propagation could be resilient to specific weight distortions, including binary weights. 

\textbf{Discussion}: the accuracy of the binary nets is significantly lowered when dealing with large CNNs such as GoogleNet. Another drawback of such binary nets is that existing binarization schemes are based on simple matrix approximations and ignore the effect of binarization on the accuracy loss. To address this issue, the work in \cite{DBLP:journals/corr/HouYK16} proposed a proximal Newton algorithm with diagonal Hessian approximation that directly minimizes the loss with respect to the binary weights. The work in \cite{DBLP:journals/corr/LinCMB15} reduced the time on float point multiplication in the training stage by stochastically binarizing weights and converting multiplications in the hidden state computation to significant changes. Zhao \textit{et al.} \cite{halfwave} proposed half-wave Gaussian Quantization to learning low precision networks, achieving promissing results. 

\subsection{Network Pruning}
An early approach to pruning was the Biased Weight Decay \cite{NIPS1988_156}. The Optimal Brain Damage \cite{Cun:1990:OBD:109230.109298} and the Optimal Brain Surgeon \cite{Hassibi93secondorder} methods reduced the number of connections based on the Hessian of the loss function. Their work suggested that such pruning gave higher accuracy than magnitude-based pruning, e.g., weight decay method.  

A following trend in this direction is to prune redundant, non-informative weights in a pre-trained DNN model. For example, Srinivas and Babu \cite{DBLP:conf/bmvc/SrinivasB15} explored the redundancy among neurons, and proposed a data-free pruning method to remove redundant neurons. Han \textit{et al.} \cite{Han:2015:LBW:2969239.2969366} proposed to reduce the total number of parameters and operations in the entire network. Chen \textit{et al.} \cite{icml2015_chenc15} proposed a HashedNets model that used a low-cost hash function to group weights into hash buckets for parameter sharing. The deep compression method in \cite{han2015deep_compression} removed the redundant connections and quantized the weights, and then used Huffman coding to encode the quantized weights. In \cite{DBLP:journals/corr/UllrichMW17}, a simple regularization method based on soft weight-sharing was proposed, which included both quantization and pruning in one simple (re-)training procedure. The above pruning schemes typically produce connections pruning in DNNs.

There is also growing interest in training compact DNNs with sparsity constraints. Those sparsity constraints are typically introduced in the optimization problem as $l_0$ or $l_1$-norm regularizers. The work in \cite{DBLP:conf/cvpr/LebedevL16} imposed group sparsity constraint on the convolutional filters to achieve structured brain Damage, i.e., pruning entries of the convolution kernels in a group-wise fashion. In \cite{Zhou2016}, a group-sparse regularizer on neurons was introduced during the training stage to learn compact CNNs with reduced filters. Wen \textit{et al.} \cite{NIPS2016_6504} added a structured sparsity regularizer on each layer to reduce trivial filters, channels or even layers. In the filter-level pruning, all the above works used $l_{1}$ or $l_{2}$-norm regularizers. The work in \cite{DBLP:journals/corr/LiKDSG16} used $l_1$-norm to select and prune unimportant filters. 

\textbf{Discussion}: there are some issues of using network pruning. First, pruning with $l_1$ or $l_2$ regularization requires more iterations to converge than general methods. In addition, all pruning criteria require manual setup of sensitivity for layers, which demands fine-tuning of the parameters and could be cumbersome for some applications. Finally, network pruning usually is able to reduce model size but not improve the efficiency (training or inference time). 

\subsection{Designing Structural Matrix}
In architectures that contain fully-connected layers, it is critical to explore this redundancy of parameters in fully-connected layers, which is often the bottleneck in terms of memory consumption. These network layers use the nonlinear transforms $f(\mathbf{x},\mathbf{M}) = \sigma(\mathbf{M}\mathbf{x})$, where $\sigma(\cdot)$ is an element-wise
nonlinear operator, $\mathbf{x}$ is the input vector, and $\mathbf{M}$ is the $m \times n$ matrix of parameters \cite{NIPS2015_5869}. When $\mathbf{M}$ is a large general dense matrix, the cost of storing $mn$ parameters and computing matrix-vector products in $O(mn)$ time. Thus, an intuitive way to prune parameters is to impose $\mathbf{x}$ as a parameterized structural matrix. An $m \times n$ matrix that can be described using much fewer parameters than $mn$ 
is called a structured matrix. Typically, the structure should not only reduce the memory cost, but also dramatically accelerate the inference and training stage via fast matrix-vector multiplication and gradient computations.    

Following this direction, the work in \cite{chang2015exploration,DBLP:journals/corr/ChengYFKCC15} proposed a simple and efficient approach based on circulant projections, while maintaining competitive error rates. Given a vector $\mathbf{r}  = (r_0, r_1, \cdots,  r_{d-1})$, a circulant matrix $\mathbf{R} \in \mathbf{R}^{d \times d}$ is defined as: 
\begin{small}
\begin{align}
\mathbf{R} = \circR(\mathbf{r}) :=
\begin{bmatrix}
r_0     & r_{d-1} & \dots  & r_{2} & r_{1}  \\
r_{1} & r_0    & r_{d-1} &         & r_{2}  \\
\vdots  & r_{1}& r_0    & \ddots  & \vdots   \\
r_{d-2}  &        & \ddots & \ddots  & r_{d-1}   \\
r_{d-1}  & r_{d-2} & \dots  & r_{1} & r_{0}
\end{bmatrix}.
\label{eq:cir}
\end{align}
\end{small}
thus the memory cost becomes $\mathcal{O}(d)$ instead of $\mathcal{O}(d^{2})$. This circulant structure also enables the use of Fast Fourier Transform (FFT) to speed up the computation. Given a $d$-dimensional vector $\mathbf{r}$, the above 1-layer circulant neural network in Eq. 1 has time complexity of $\mathcal{O}(d \log d)$. 

In \cite{Yang2015}, a novel Adaptive Fastfood transform was introduced to reparameterize the matrix-vector multiplication of fully connected layers. The Adaptive Fastfood transform matrix $\mathbf{R} \in \mathbf{R}^{n \times d}$ was defined as:  
\begin{equation}
\mathbf{R} = \mathbf{S}\mathbf{H}\mathbf{G}\mathbf{\Pi}\mathbf{H}\mathbf{B}
\label{eq:fastfodd}
\end{equation}
where $\mathbf{S}$, $\mathbf{G}$ and $\mathbf{B}$ are random diagonal matrices. $\mathbf{\Pi} \in \{0,1\}^{d \times d}$ is a random permutation matrix, and $\mathbf{H}$ denotes the Walsh-Hadamard matrix. Reparameterizing a fully connected layer with $d$ inputs and $n$ outputs using the Adaptive Fastfood transform reduces the storage and the computational costs from $\mathcal{O}(nd)$ to $\mathcal{O}(n)$ and from $\mathcal{O}(nd)$ to $\mathcal{O}(n \log d)$, respectively.

The work in \cite{NIPS2015_5869} showed the effectiveness of the new notion of parsimony in the theory of structured matrices. Their proposed method can be extended to various other structured matrix classes, including block and multi-level Toeplitz-like \cite{Chun1991} matrices related to multi-dimensional convolution \cite{DBLP:journals/siamsc/RakhubaO15}. Following this idea, \cite{Moczulski2015} proposed a general structured efficient linear layer for CNNs. 

\textbf{Drawbacks}: one issue of this kind of approaches is that the structural constraint usually hurts the performance since the constraint might bring bias to the model. On the other hand, how to find a proper structural matrix is hard. There is no theoretical way to derive it out. 

\section{Low-rank Approximation and Sparsity}
Convolution operations contribute the bulk of most computations in deep DNNs, thus reducing the convolution layer would improve the compression rate as well as the overall speedup. The convolution kernels can be viewed as a 3D tensor. Ideas based on tensor decomposition is derived by the intuition that there is a structure spacity in the 3D tensor. Regarding the fully-connected layer, it can be view as a 2D matrix (or 3D tensor) and the low-rankness can also help. 

\begin{figure}
\centering
\includegraphics[width=8cm]{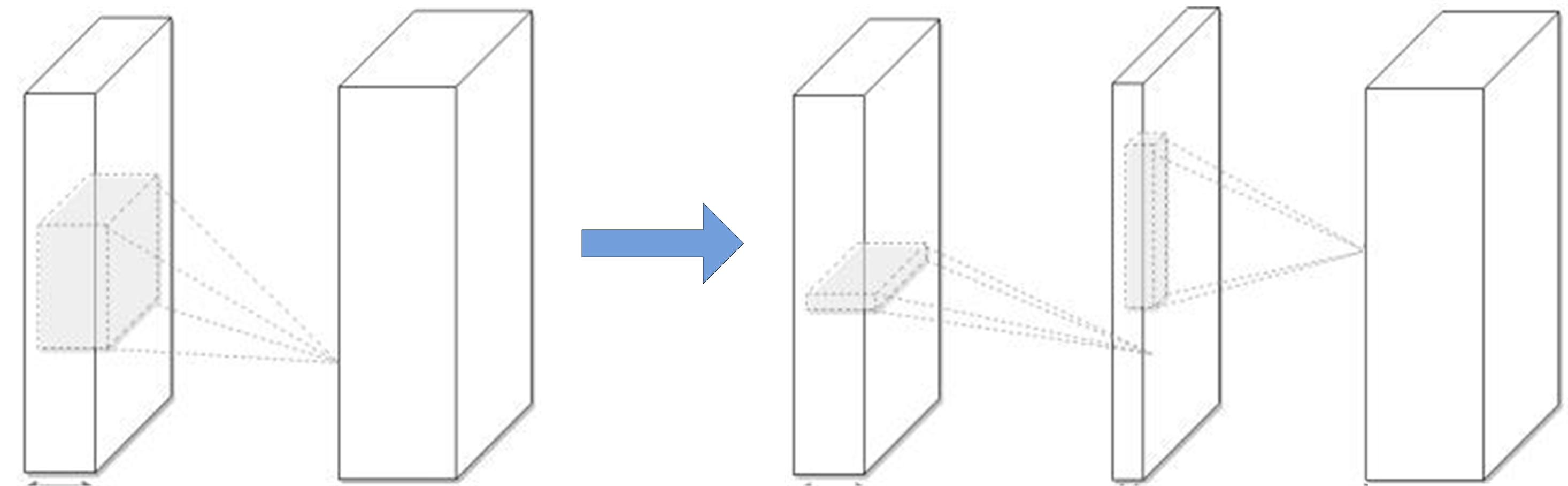}
\caption{A typical framework of the low-rank regularization method. The left is the original convolutional layer and the right is the low-rank constraint convolutional layer with rank-K.}
\label{fig:lowrank}
\end{figure}
It has been a long time for using low-rank filters to accelerate convolution, for example, high dimensional DCT (discrete cosine transform) and wavelet systems using tensor products to be constructed from 1D DCT transform and 1D wavelets respectively. Learning separable 1D filters was introduced by Rigamonti \textit{et al.} \cite{DBLP:conf/cvpr/RigamontiSLF13} using a dictionary learning approach. For some simple DNN models, a few low-rank approximation and clustering schemes for the convolutional kernels were proposed in \cite{NIPS2014_5544}. They achieved 2$\times$ speedup for a single convolutional layer with 1\% drop in classification accuracy. The work in \cite{BMVC_2014} proposed to use different tensor decomposition schemes, reporting a 4.5$\times$ speedup with 1\% drop in accuracy in text recognition. 

The low-rank approximation was done layer by layer. The parameters of one layer were fixed after it was done, and the layers above were fine-tuned based on a reconstruction error criterion. These are typical low-rank methods for compressing 3D convolutional layers, which is described in Figure \ref{fig:lowrank}. Following this direction, Canonical Polyadic (CP) decomposition of was proposed for the kernel tensors in \cite{DBLP:journals/corr/LebedevGROL14}. Their work used nonlinear least squares to compute the CP decomposition.  In \cite{DBLP:journals/corr/TaiXWE15}, a new algorithm for computing the low-rank tensor decomposition for training low-rank constrained CNNs from scratch were proposed. It used Batch Normalization (BN) to transform the activation of the internal hidden units. In general, both the CP and the BN decomposition schemes in \cite{DBLP:journals/corr/TaiXWE15} (BN Low-rank) can be used to train CNNs from scratch. However, there are few differences between them. For example, finding the best low-rank approximation in CP decomposition is an ill-posed problem, and the best rank-$K$ ($K$ is the rank number) approximation may not exist sometimes. While for the BN scheme, the decomposition always exists. We perform a simple comparison of both methods shown in Table \ref{tab:lowrank:res}. The actual speedup and the compression rates are used to measure their performances.

\begin{table}[h]
\footnotesize
\caption{Comparisons between different low-rank models and their baselines on ILSVRC-2012. }
\label{tab:lowrank:res}
\centering
\begin{tabular}{cccc}
Model & TOP-5 Accuracy & Speed-up & Compression Rate \\
\hline AlexNet & 80.03\% & 1. & 1. \\ 
BN Low-rank & 80.56\% & 1.09 & 4.94 \\
CP Low-rank & 79.66\% & 1.82 & 5. \\
\hline VGG-16 & 90.60\% & 1. & 1.  \\
BN Low-rank & 90.47\% & 1.53 & 2.72 \\
CP Low-rank & 90.31\% & 2.05 & 2.75 \\
\hline GoogleNet & 92.21\% & 1. & 1. \\
BN Low-rank & 91.88\% & 1.08 & 2.79 \\
CP Low-rank & 91.79\% & 1.20 & 2.84 \\
\end{tabular}
\end{table}

As we mentioned before, the fully connected layers can be viewed as a 2D matrix and thus the above mentioned methods can also be applied there. There are several classical works on exploiting low-rankness in fully connected layers. For instance, Misha \textit{et al.} \cite{NIPS2013_5025} reduced the number of dynamic parameters in deep models using the low-rank method. \cite{Sainath13low-rankmatrix} explored a low-rank matrix factorization of the final weight layer in a DNN for acoustic modeling. In \cite{DBLP:journals/corr/LuKZCJF16}, Lu \textit{et al.} adopted truncated SVD (singular value decomposition) to decompose the fully connected layer for designing compact multi-task deep
learning architectures.

\textbf{Discussion}: low-rank approximation based approaches are straightforward for model compression and acceleration. However, the implementation is not that easy since it involves decomposition operation, which is computationally expensive. Another issue is that current methods perform low-rank approximation layer by layer, thus cannot perform global parameters compression, which is important as different layers hold different information. Finally, factorization requires extensive model retraining to achieve convergence when compared to the original model. 

\section{Transferred/compact convolutional filters}
CNNs are parameter efficient due to exploring the translation invariant property of the representations to the input image, which is the key to the success of training very deep models without severe over-fitting. Although a strong theory is currently missing, a large number of empirical evidence support the notion that both the translation invariant property and the convolutional weight sharing are important for good predictive performance. The idea of using transferred convolutional filters to compress CNN models is motivated by recent works in \cite{cohen2016group}, which introduced the equivariant group theory. Let $\mathbf{x}$ be an input,  $\Phi(\cdot)$ be a network
or layer and $\mathcal{T}(\cdot)$ be the transform matrix. The concept of equivalence is defined as:

\begin{equation}
\mathcal{T}^{`} \Phi(\mathbf{x}) = \Phi(\mathcal{T}\mathbf{x})
\label{eq:intro}
\end{equation}
indicating that transforming the input $\mathbf{x}$ by the transform $\mathcal{T}(\cdot)$ and then passing it through the network or layer $\Phi(\cdot)$ should give the same result as first mapping $\mathbf{x}$ through the network and then transforming the representation. Note that in Eq. (\ref{eq:intro}), the transforms $\mathcal{T}(\cdot)$ and $\mathcal{T}^{'}(\cdot)$ are not necessarily the same as they operate on different objects. According to this theory, it is reasonable applying transform to layers or filters $\Phi(\cdot)$ to compress the whole network models. From empirical observation, deep CNNs also benefit from using a large set of convolutional filters by applying certain transform $\mathcal{T}(\cdot)$ to a small set of base filters since it acts as a regularizer for the model. 

Following this direction, there are many recent reworks proposed to build a convolutional layer from a set of base filters\cite{doublecnn,crelu,mba,cohen2016group}. What they have in common is that the transform $\mathcal{T}(\cdot)$ lies in the family of functions that only operate in the spatial domain of the convolutional filters. For example, the work in \cite{crelu} found that the lower convolution layers of CNNs learned redundant filters to extract both positive and negative phase information of an input signal, and defined $\mathcal{T}(\cdot)$ to be the simple negation function:
\begin{equation}
\mathcal{T}(\mathbf{W}_{x}) = \mathbf{W}^{-}_{x}
\label{eq:intro}
\end{equation}
where $\mathbf{W}_{x}$ is the basis convolutional filter and $\mathbf{W}^{-}_{x}$ is the filter consisting of the shifts whose activation is opposite to that of $\mathbf{W}_{x}$ and selected after max-pooling operation. By doing this, the work in \cite{crelu} can easily achieve 2$\times$ compression rate on all the convolutional layers. It is also shown that the negation transform acts as a strong regularizer to improve the classification accuracy. The intuition is that the learning algorithm with pair-wise positive-negative constraint can lead to useful convolutional filters instead of redundant ones.  

In \cite{mba}, it was observed that magnitudes of the responses from convolutional kernels had a wide diversity of pattern representations in the network, and it was not proper to discard weaker signals with a single threshold. Thus a multi-bias non-linearity activation function was proposed to generates more patterns in the feature space at low computational cost. The transform $\mathcal{T}(\cdot)$ was define as:
\begin{equation}
\mathcal{T}^{`} \Phi(\mathbf{x}) = \mathbf{W}_{x} + \delta
\label{eq:intro}
\end{equation}
\noindent where $\delta$ were the multi-bias factors. The work in \cite{Dieleman:2016:ECS:3045390.3045590} considered a combination of rotation by a multiple of $90^{\circ}$ and horizontal/vertical flipping with:
\begin{equation}
\mathcal{T}^{`} \Phi(\mathbf{x}) = \mathbf{W}^{T_{\theta}}
\label{eq:intro}
\end{equation}
\noindent where $\mathbf{W}^{T_{\theta}}$ was the transformation matrix which rotated the original filters with angle 
$\theta \in \{ 90,180,270 \}$. In \cite{cohen2016group}, the transform was generalized to any angle learned from data, and $\theta$ was directly obtained from data. Both works \cite{Dieleman:2016:ECS:3045390.3045590} and \cite{cohen2016group} can achieve good classification performance. 

The work in \cite{doublecnn} defined $\mathcal{T}(\cdot)$ as the set of translation functions applied to 2D filters:
\begin{equation} \label{eq:transcorrelation}
\mathcal{T}^{`} \Phi(\mathbf{x}) = T(\cdot, x, y)_{x, y \in \{-k, ..., k\},  (x,y) \neq (0, 0)} 
\end{equation}
\noindent where $T(\cdot, x, y)$ denoted the translation of the first operand by $(x,y)$ along its spatial dimensions, with proper zero padding at borders to maintain the shape. The proposed framework can be used to 1) improve the classification accuracy as a regularized version of maxout networks, and 2) to achieve parameter efficiency by flexibly varying their architectures to compress networks.  

Table \ref{tab:transfer:res} briefly compares the performance of different methods with transferred convolutional filters, using VGGNet (16 layers) as the baseline model. The results are reported on CIFAR-10 and CIFAR-100 datasets with Top-5 error. It is observed that they can achieve reduction in parameters with little or no drop in classification accuracy. 

\begin{table}[h]
\footnotesize
\caption{A simple comparison of different approaches on CIFAR-10 and CIFAR-100.}
\label{tab:transfer:res}
\centering
\begin{tabular}{cccc}
Model & CIFAR-100 & CIFAR-10 & Compression Rate \\
\hline VGG-16 & 34.26\% & 9.85\% & 1. \\ 
MBA \cite{mba} & 33.66\% & 9.76\% & 2.  \\
CRELU \cite{crelu} & 34.57\% & 9.92\% & 2.  \\
CIRC \cite{cohen2016group} & 35.15\% & 10.23\% & 4.  \\
DCNN \cite{doublecnn} & 33.57\% & 9.65\% & 1.62  \\
\end{tabular}
\end{table}

\textbf{Discussions}: there are a few issues to be addressed for approaches that apply transform constraints to convolutional filters. First, these methods can achieve competitive performance for wide/flat architectures (e.g., VGGNet, AlexNet) but not thin/deep ones (e.g., ResNet). Secondly, the transfer assumptions sometimes are too strong to guide the learning, making the results unstable in some situation.

Using a compact filter for convolution can directly reduce the computation cost. The key idea is to replace the loose and over-parametric filters with compact blocks to improve the speed. Decomposing $3 \times 3$ convolution into two $1 \times 1$ convolutions was used in \cite{journals/corr/SzegedyIV16}, which achieved significant acceleration. SqueezeNet \cite{DBLP:journals/corr/WuIJK16} was proposed to replace $3 \times 3$ convolution with $1 \times 1$ convolution, which created a compact neural network with about 50 fewer parameters. Similar technique has been adapted in MobileNets \cite{mobilenets}.

\section{Knowledge Distillation}
To the best of our knowledge, exploiting knowledge transfer (KT) to compress model was first proposed by Caruana \textit{et al.} \cite{Bucilua:2006:MC:1150402.1150464}. They trained a compressed/ensemble model of strong classifiers with pseudo-data labeled, and reproduced the output of the original larger network. But the work is limited to shallow models. The idea has been recently adopted in \cite{DBLP:conf/nips/BaC14} as knowledge distillation (KD) to compress deep and wide networks into shallower ones, where the compressed model mimicked the function learned by the complex model. The main idea of KD based approaches is to shift knowledge from a large teacher model into a small one by learning the class distributions output via softmax. 

The work in \cite{DBLP:journals/corr/HintonVD15} introduced a KD compression framework, which eased the training of deep networks by following a student-teacher paradigm, in which the student was penalized according to a softened version of the teacher's output. The framework compressed an ensemble of teacher networks into a student network of similar depth. The student was trained to predict the output and the classification labels. Despite its simplicity, KD demonstrates promising results in various image classification tasks. The work in \cite{DBLP:journals/corr/RomeroBKCGB14} aimed to address the network compression problem by taking advantage of depth neural networks. It proposed an approach to train thin but deep networks, called FitNets, to compress wide and shallower (but still deep) networks. The method was extended the idea to allow for thinner and deeper student models. In order to learn from the intermediate representations of teacher network, FitNet made the student mimic the full feature maps of the teacher. However, such assumptions are too strict since the capacities of teacher and student may differ greatly. 

All the above approaches are validated on MNIST, CIFAR-10, CIFAR-100, SVHN and AFLW benchmark datasets, and experimental results show that these methods match or outperform the teacher's performance, while requiring notably fewer parameters and multiplications. 

There are several extension along this direction of distillation knowledge. The work in \cite{NIPS2015_5965} trained a parametric student model to approximate a Monte Carlo teacher. The proposed framework used online training, and used deep neural networks for the student model. Different from previous works which represented the knowledge  using the soften label probabilities, \cite{DBLP:conf/aaai/LuoZLWT16} represented the knowledge by using the neurons in the higher hidden layer, which preserved as much information as the label probabilities, but are more compact. The work in \cite{DBLP:journals/corr/ChenGS15} accelerated the experimentation process by instantaneously transferring the knowledge from a previous network to each new deeper or wider network. The techniques are based on the concept of function-preserving transformations between neural network specifications. Zagoruyko \textit{et al.} \cite{DBLP:journals/corr/ZagoruykoK16a} proposed Attention Transfer (AT) to relax the assumption of FitNet. They transferred the attention maps that are summaries of the full activations.

\textbf{Discussions}: KD-based approaches can make deeper models shallower and help significantly reducing the computational cost. However, there are a few disadvantages. One of those is that KD can only be applied to tasks with softmax loss function, which hinders its usage. Another drawback is that KD-based approaches generally achieve less competitive performance compared with other type of approaches. 

\section{Other Types of Approaches}
We first summarize the works utilizing attention-like mechanism \cite{attention}, which can reduce computations significantly by learning to selectively focus or ``attend'' to a few, task-relevant input regions. In \cite{DBLP:journals/pami/WuPKLSDO16}, dynamic deep neural networks (D2NN) were introduced, which were a type of feed-forward deep neural network that selected and executed a subset of D2NN neurons based on the input. The dynamic capacity network (DCN) \cite{DBLP:conf/icml/AlmahairiBCZLC16} that combined the small sub-networks with low capacity, and the large ones with high capacity. The attention mechanism was used to direct the high-capacity sub-networks to focus on the task-relevant regions. By dong this, the size of the model has been significantly reduced. Following this direction, the work in \cite{45929} introduced the conditional computation idea, which only computes the gradient for some important neurons via a sparsely-gated mixture-of-experts Layer (MoE).

There have been other attempts to reduce the number of parameters of neural networks by replacing the fully connected layer with global average pooling \cite{43022,doublecnn}. Network architecture such as GoogleNet or Network in Network, can achieve state-of-the-art results on several benchmarks by adopting this idea. However, these architectures have not been fully optimized the utilization of the computing resources inside the network. This problem was noted by Szegedy \textit{et al.} \cite{43022} and motivated them to increase the depth and width of the network while keeping the computational budget constant.

The work in \cite{Huang2016} targeted the Residual Network based model with a spatially varying computation time, called stochastic depth, which enabled the seemingly contradictory setup to train short networks and used deep networks at test time. It started with very deep networks, while during training, for each mini-batch, randomly dropped a subset of layers and bypassed them with the identity function. Following this direction, thew work in \cite{DBLP:journals/corr/YamadaIK16} proposed a pyramidal residual networks with stochastic depth. In \cite{blockdrop}, Wu \textit{et al.} proposed an approach that learns to dynamically choose which layers of a deep network to execute during inference so as to best reduce total computation. Veit \textit{et al.} exploited convolutional networks with adaptive inference graphs to adaptively define their network topology conditioned on the input image \cite{Veit2018}.

Other approaches to reduce the convolutional overheads include using FFT based convolutions \cite{b70224ae04784e15b91d2056c46924a6} and fast convolution using the Winograd algorithm \cite{DBLP:conf/cvpr/LavinG16}. 
Zhai \textit{et al.} \cite{DBLP:journals/corr/ZhaiWKCLZF16} proposed a strategy call stochastic spatial sampling pooling, which speed-up the pooling operations by a more general stochastic version. Saeedan \textit{et al.} presented a novel pooling layer for convolutional
neural networks termed detail-preserving pooling (DPP),
based on the idea of inverse bilateral filters \cite{saeedan2018dpp}. Those works only aim to speed up the computation but not reduce the memory storage. The MobileNetV2 \cite{Sandler_2018_CVPR} proposed the novel inverted residual structure. 

\section{Benchmarks, Evaluation and Databases}
In the past years the deep learning community had made great efforts in benchmark. One of the most well-known model used in compression and acceleration for CNNs is Alexnet \cite{krizhevsky2012}, which has been occasionally used for assessing the performance of compression. Other popular standard models include LeNets \cite{Lecun98gradient-basedlearning}, All-CNN-nets \cite{DBLP:journals/corr/SpringenbergDBR14} and many others. LeNet-300-100 is a fully connected network with two hidden layers, with 300 and 100 neurons each. LeNet-5 is a convolutional network that has two convolutional layers and two fully connected layers. Recently, more and more state-of-the-art architectures are used as baseline models in many works, including network in networks (NIN) \cite{chen2014}, VGG nets \cite{Simonyan14c} and  residual networks (ResNet) \cite{He2015}. Table \ref{tab:baseline} summarizes the baseline models commonly used in several typical compression methods. 

\begin{table}
\caption{Summarization of baseline models used in different representative works of network compression.}
\label{tab:baseline}
\centering
\begin{tabular}{c|c}
\hline Baseline Models & Representative Works \\
\hline Alexnet \cite{krizhevsky2012} &  structural matrix \cite{Yang2015,chang2015exploration,NIPS2015_5869}  \\
& low-rank factorization \cite{DBLP:journals/corr/TaiXWE15} \\
\hline Network in network \cite{chen2014} &  low-rank factorization \cite{DBLP:journals/corr/TaiXWE15} \\
\hline VGG nets \cite{Simonyan14c} & transferred filters \cite{doublecnn} \\
&  low-rank factorization \cite{DBLP:journals/corr/TaiXWE15} \\
\hline Residual networks \cite{He2015} & compact filters \cite{DBLP:journals/corr/WuIJK16}, stochastic depth \cite{Huang2016}  \\
& parameter sharing \cite{DBLP:journals/corr/UllrichMW17}\\
\hline All-CNN-nets \cite{DBLP:journals/corr/SpringenbergDBR14} & transferred filters \cite{crelu} \\ 
\hline LeNets \cite{Lecun98gradient-basedlearning} & parameter sharing \cite{DBLP:journals/corr/UllrichMW17} \\ 
& parameter pruning \cite{Han:2015:LBW:2969239.2969366,Hassibi93secondorder} \\
\hline
\end{tabular}
\end{table}

The standard criteria to measure the quality of model compression and acceleration are the compression and the speedup rates. Assume that $a$ is the number of the parameters in the original model $M$ and $a^{*}$ is that of the compressed model $M^{*}$, then the compression rate $\alpha(M,M^{*})$ of $M^{*}$ over $M$ is:
\begin{equation}
\alpha(M,M^{*}) = \frac{a}{a^{*}}.
\label{eq:intro}
\end{equation}
Another widely used measurement is the index space saving defined in several papers \cite{chang2015exploration,Moczulski2015} as
\begin{equation}
\beta(M,M^{*}) = \frac{a-a^{*}}{a^{*}},
\label{eq:intro}
\end{equation}
where $\beta(M,M^{*})$ is the defined space saving rate.

Similarly, given the running time $s$ of $M$ and $s^{*}$ of $M^{*}$, the speedup rate $\delta(M,M^{*})$ is defined as:
\begin{equation}
\delta(M,M^{*}) = \frac{s}{s^{*}}.
\label{eq:intro}
\end{equation}
Most work used the average training time per epoch to measure the running time, while in \cite{chang2015exploration,Moczulski2015}, the average testing time  was used. Generally, the compression rate and speedup rate are highly correlated, as smaller models often results in faster computation for both the training and the testing stages. 

A good compression method is expected to achieve almost the same performance as the original model with much smaller parameters and less computational time. However, for different applications with different CNN designs, the relation between parameter size and computational time might be different. For example, it is observed that for deep CNNs with fully connected layers, most of the parameters are in the fully connected layers; while for image classification tasks, float point operations are mainly in the first few convolutional layers since each filter is convolved with the whole image, which is usually very large at the beginning. Thus compression and acceleration of the network should focus on different type of layers for different applications.

\section{Challenges and Future Work}
We summarized recent efforts on compressing and accelerating deep neural networks (DNNs). Here we discuss more details about how to choose different compression approaches, technique challenges and possible solutions for future work.

\subsection{General Suggestions}
There are no golden criteria to measure which approach is the best. How to choose a proper method really depends on the applications and requirements. Here are some general suggestions we can provide: 
\begin{itemize}
\item If the applications need compacted models from pre-trained deep nets, you can choose either pruning \& quantization or low rank factorization based methods. If you need end-to-end solutions for your problem, the low rank and transferred convolutional filters approaches should be considered. 

\item For applications in particular domains (e.g., medical images), methods with human prior (like the transferred convolutional filters, structural matrix) sometimes have benefits. For example, when doing medical images classification, transferred convolutional filters could work well as medical images (like organ) do have the rotation transformation property.   
\item The approaches of pruning \& quantization generally give reasonable compression rate while not hurt the accuracy. Thus for applications which requires stable model performance, it is better to utilize pruning \& quantization.

\item If your application involves small/medium size datasets or requires significantly improving efficiency, you can try the knowledge distillation approaches. The compressed student model can take the benefit of transferring knowledge from teacher model, achieving robust performance when datasets are not large.   

\item As we mentioned before, these aforementioned techniques are orthogonal. It is reasonable to combine two or three of them to maximize the gain. For some specific applications, like object detection, which requires both convolutional and fully connected layers, you can compress the convolutional layers with a low rank based method and the fully connected layers with a pruning technique. 
\end{itemize}

\subsection{Technique Challenges}
We also summarize the following challenges still need to be addressed. 
\begin{itemize}
\item Most of the current state-of-the-art approaches build on well-designed CNN models, which have limited freedom to change the configuration (e.g., network architectures, hyper-parameters). To handle more complicated tasks, the furture work should provide more plausible ways to configure the compressed models. 

\item Hardware constraints in various of small platforms (e.g., mobile, robotic, self-driving car) are still a major problem to hinder the extension of deep CNNs. How to make full use of the limited computational source and how to design special compression methods for such platforms are still challenges that need to be addressed. 

\item Pruning is an effective way to compress and accelerate CNNs. The current pruning techniques are mostly designed to eliminate connections between neurons. On the other hand, pruning channel can directly reduce the feature map width and shrink the model into a thinner one. It is efficient but also challenging because removing channels might dramatically change the input of the following layer.

\item As we mentioned before, methods of structural matrix and transferred convolutional filters impose prior human knowledge to the model, which could significantly affect the performance and stability. It is critical to investigate how to control the impact of those prior knowledge. 

\item The methods of knowledge distillation provide many benefits such as directly accelerating model without special hardware or implementations. It is still worthy developing KD-based approaches and exploring how to improve their performances. 

\item Despite the great achievements of these compression approaches, the black box mechanism is still the key barrier to the adoption. For example, why some neurons/connections are pruned is not clear. Exploring the knowledge interpret-ability is still an important challenge. 
\end{itemize}

\subsection{Possible Future Directions}
To solve the hyper-parameters configuration problem, we can rely on the recent neural architecture search strategies \cite{45826,liu2018darts}. This framework provides a mechanism allowing the algorithm to automatically learn how to exploit structure in the problem of interest. Leveraging reinforcement learning to efficiently sample the design space and improve the model compression has been tried in \cite{He_2018_ECCV}. 

Regarding the use of CNNs in different hardware platforms, proposing some hardware-aware approaches is one direction.
Wang \textit{et al.} \cite{haq} proposed the Hardware-Aware Automated Quantization (HAQ) to take the hardware accelerator`s feedback in the design loop. Similar idea can be applied to make CNNs more applicable for different platforms. The work in \cite{cai2018proxylessnas} directly learn the architectures for large-scale target tasks and target hardware based performance. 

Channel pruning provides the efficiency benefit on both CPU and GPU because no special implementation is required. But it is also challenging to handle the input configuration. One possible solution is to use the training-based channel pruning methods \cite{NIPS2016_6372}, which focus on imposing sparse constraints on weights during training. In addition, training from scratch for such methods is costly for deep CNNs. In \cite{He_2017_ICCV}, the authors provided an iterative two-step algorithm to effectively prune channels in each layer. The work in \cite{Liu2017learning} associated a scaling factor with each channel and imposed regularization on these scaling factors during training to automatically identify unimportant channels. Liu \textit{et al.} \cite{liu2018rethinking} showed that pruned architecture itself is more crucial and pruning can be useful as an architecture search paradigm. 

Exploring new types of knowledge in the teacher models and transferring it to the student models is useful for the knowledge distillation (KD) approaches. Instead of directly reducing and transferring parameters, passing selectivity knowledge of neurons could be helpful. One option is deriving a way to select essential neurons related to the task \cite{SSS2018,DBLP:conf/aaai/ChenWZ18}. Very recently, the contrastive loss instead of KL divergence for distillation has been tried in \cite{Tian2020Contrastive}. 

For methods with the convolutional filters and the structural matrix, we can conclude that the transformation lies in the family of functions that only operations on the spatial dimensions. Hence to address the imposed prior issue, one solution is to provide a generalization of the aforementioned approaches in two aspects: 1) instead of limiting the transformation to belong to a set of predefined transformations, let it be the whole family of spatial transformations applied on 2D filters or matrix, and 2) learn the transformation jointly with all the model parameters.  

Despite the image classification task, people are also adapting the compacted models in other tasks \cite{NIPS2017_6676,Sandler_2018_CVPR,DBLP:conf/cvpr/HuangRSZKFFWSG017}. There is also some work about deep natural language models \cite{suncheng2019,DistilBERT}.  We would like to see more work for applications with larger deep nets (e.g., video and image frames \cite{Cheng_2014_CVPR,Cao2012IBMRA}, vision + language \cite{chen2019uniter} and GANs \cite{mroueh2018sobolev,gan}). 

\section{ACKNOWLEDGMENTS}
The authors would like to thank the reviewers and broader community for their feedback on this survey. In particular, we would like to thank Hong Zhao from the Department of Automation of Tsinghua University for her help on modifying the paper. 

\bibliographystyle{IEEEtran}
\bibliography{IEEEabrv,reference}\ 

\end{document}